\documentclass{article}

\usepackage{graphicx} 
\usepackage[utf8]{inputenc}
\usepackage{amsmath,amssymb}
\usepackage{graphicx}
\usepackage{hyperref}
\usepackage{natbib}
\usepackage{geometry}
\title{Semiotic Complexity and Its Epistemological Implications for Modeling Culture}

\author{
    Zachary K. Stine\thanks{Corresponding author: \texttt{zstine@uca.edu}} \\
    Department of Computer Science and Engineering \\ 
    University of Central Arkansas \\
    \and 
    James E. Deitrick\\
    Department of Philosophy and Religious Studies \\ 
    University of Central Arkansas
}

\begin{document}

\maketitle

\begin{abstract}
  Greater theorizing of methods in the computational humanities is needed for epistemological and interpretive clarity, and therefore the maturation of the field. In this paper, we frame such modeling work as engaging in translation work from a cultural, linguistic domain into a computational, mathematical domain, and back again. Translators benefit from articulating the theory of their translation process, and so do computational humanists in their work---to ensure internal consistency, avoid subtle yet consequential translation errors, and facilitate interpretive transparency. Our contribution in this paper is to lay out a particularly consequential dimension of the lack of theorizing and the sorts of translation errors that emerge in our modeling practices as a result. Along these lines we introduce the idea of semiotic complexity as the degree to which the meaning of some text may vary across interpretive lenses, and make the case that dominant modeling practices--especially around evaluation--commit a translation error by treating semiotically complex data as semiotically simple when it seems epistemologically convenient by conferring superficial clarity. We then lay out several recommendations for researchers to better account for these epistemological issues in their own work.
\end{abstract}

\section{Introduction}

The use of computational methods in the study of cultural artifacts---from models like linear regression and artificial neural networks, to how we evaluate and interpret those models---can be usefully understood as a kind of translation work from a complex, cultural medium into a formal, computational medium. Research questions arise in the cultural domain within culturally-embedded minds. When a researcher designs a computational model to aid in answering such a question, they translate from the cultural into the computational in each modeling decision they make. After completing this first translation problem, the researcher then makes use of the model by interpreting it (either directly or in downstream outputs that depend on it), requiring a second translation to be made, now from the computational going back into the cultural, by way of culturally-embedded researchers making sense of them.

In these bidirectional translation problems, we as researchers want to ensure that our translations are reasonable, that they can be sufficiently evaluated and understood by others engaged in collective knowledge-building. Yet translation work can vary in the complexity required to interpret and evaluate it. 

Consider, for example, how evaluating a translation of ``hello" into modern Mandarin Chinese is much simpler than evaluating a translation of a text from classical (i.e., literary) Chinese, like the \textit{Zhuangzi}, into modern English. On the surface, both translation tasks may look alike. They both involve translating meaningful symbols from one orthographically distinct language into another. Yet the latter translation task differs significantly as a result of the inherent complexity of translating from classical Chinese into English \cite{rosemont_ames_2016} in addition to the internal complexity of the source text itself (see the preface to \cite{Zhuangzi_Ziporyn_2020}).

Translating a text like the \textit{Zhuangzi} necessarily involves significant interpretive choices to be made, and in order for these interpretive choices to be identified, understood, and evaluated in light of alternative decisions by others, translators ought to articulate their own theory of the text and its translation from source to target language \cite{rosemont_ames_2016}. Errors or inconsistencies in a simple translation problem ought to be more noticeable and easily diagnosed than in complex translation problems, which may give rise to complex translation errors. The more complex a translation problem, then, the more important it is to understand the translator's own assumptions and theoretical commitments. 

The same is true for modeling work undertaken in the computational humanities. The complexity of translating from the cultural to the computational is immense, owing to the distinct affordances of the two domains. Importantly, if this initial translation from cultural to computational cannot be sufficiently evaluated, then neither can our interpretations of our models. While modeling might seem to hold out the promise of breaking the recursivity of studying culture from inside culture, we cannot use modeling without first translating back into the cultural. Models, then, ultimately become cultural objects \cite{seaver_2017, Farrell_Gopnik_et_al_2025}.  

We make the following contributions in the sections below. First, we build on this metaphor of cultural modeling as translation work in section \ref{sec:translation_errors} to explore the epistemological consequences for this field that are posed by translation errors, which emerge out of theoretical misalignment between modeler and model, and which may hide within the perceived clarity of model evaluation practices. In section \ref{sec:semiotic_complexity}, we theorize that such translation errors arise when we do not account for semiotic complexity—a property that corresponds to the interpretive variance of complex cultural phenomena—which is not indicated by data or model types. We then consider starting points for accounting with semiotic complexity in section \ref{sec:recommendations}.

\section{Translation Errors from Latent Theoretical Misalignment}
\label{sec:translation_errors}

If translation errors are problematic when translating from one natural language to another, it is reasonable to assume they will also be problematic when translating research questions from the cultural domain to the computational and back again. These translation errors are especially problematic because they only become visible to us once we have a sufficiently clear theory. Consider the previous translation example in terms of evaluating a machine translation model. Justifying a translation of ``hello" will not need the same degree of theoretical complexity as what is needed in justifying a translation of the \textit{Zhuangzi}. When we translate cultural questions into computational methods, we are doing something like a translation of \textit{Zhuangzi} into mathematics.

The call for theorizing, not just in the cultural domain, but in the computational, has been longstanding \cite{underwood2014}. Yet there has been an unfortunate tendency to assume that machine learning allows for the atheoretical modeling of data, though this is not the case \cite{andrews2023immortal}. Instead, a researcher may simply be unaware of the theoretical commitments implied by models, which may cohere with different theories to different degrees. 
Misalignment between the theoretical commitments of the modeler and model enable translation errors with significant epistemological consequences, undermining both the interpretability of results and our capacity for meaningful collective knowledge-building. To evaluate or build upon a model, one must understand how it follows from the researcher’s theory, thereby making visible any tensions or gaps between the question posed and what the model actually investigates. Without this clarity, our findings risk doing more to expand vocabulary than to advance actual claims as Mallory \cite{Mallory2025} discusses with respect to agentive language around machine learning. Without theoretical orientation, we complicate our ability to understand not only the answers we obtain from our methods, but also the questions our methods actually pose. 

To illustrate how theoretical commitments can be latent within models, we consider two binary classification models of text. The first binary classifier takes a sequence of text as input and predicts if the text belongs to one of two orthographically distinct languages (e.g., English or Mandarin). The second binary classifier takes sequences of text (all in the same language) and predicts if the text is religious or not. Such binary classification models are a simple kind of supervised learning problem, and their evaluation typically proceeds by calculating performance metrics, like accuracy or $F_1$ measures, from predictions made on held-out test data. If the model performs well on such measures, then a researcher might be tempted to treat the model as permitting theory-free analysis of the categories it distinguishes between. While this may be slightly more reasonable for the first classifier, it is very much not the case for the second, as a result of the complexity of religion, which might be defined in myriad ways that are internally reasonable, yet do not cohere with each other \cite{smith1998religion}. 

The decisions underlying the religious text classifier demarcate a space of theories of religion with which the model may align. It is possible for the model to perform well on the classification task, while also being deeply inconsistent with the theoretical commitments of the researchers using it, and therefore of limited research value. For example, suppose one observation, $x_i$, consists of a description of prayers being made to a deity and this observation is labeled as religious with $y_i = 1$. Now suppose another observation, $x_j$, describes a series of rituals undertaken by a sports fan prior to their favorite team playing, which is labeled as not religious with $y_j = 0$. Such a model would thus be aligned with theories of religion that prioritize belief in deities and prayer as markers of religiosity, but which do not treat ritualistic efforts to effect change in the world, like magic, as religious. One may use such a model as a kind of sense-making exercise \cite{bamman2024classification}, but not as anything like a uniquely authoritative model of religion, supported by ground truth evaluation.

The most troubling aspect here is that theorizing is not only needed to avoid such translation errors, but is needed for such errors to become visible. Recalling G\"odel, the computational model cannot be \textit{computationally} evaluated with respect to its coherence with the cultural theory it is meant to translate, but must be evaluated exogenously. Such translation errors may even be obscured by the apparent clarity of the performance metrics used to evaluate the model. This is because, as Nguyen \cite{Nguyen_2021} argues, clarity often functions as a thought-ending heuristic. Thus, simple evaluation methods for complex cultural modeling problems risks the cultural complexity being lost in translation, flattened within the computational domain into something that bears little resemblance to the researcher's purported object of study.

Translation errors are not a new problem for the humanities. For example, comparative religion research has long grappled with the problem of unstated cultural assumptions detracting from the usefulness of the work \cite{paden2009comparative}. Humanities scholars cannot use computers to escape from having to give an account for our assumptions and must attempt to articulate them so that they do not confound us in hiding.

\section{Semiotic Complexity}
\label{sec:semiotic_complexity}

In the two illustrations we have described---language translation and binary classification---we have shown that modeling problems may appear to be identical, yet differ in significant ways, particularly with respect to the importance of exogenous evaluation, which requires theoretical clarity about the relationship between a research question that arises in the cultural domain and its purported translation in the computational domain. Our contribution in this section is to define and theorize this difference as resulting from semiotic complexity. We argue that semiotic complexity, while not reducible to data type or modeling task, should be taken seriously as a property of cultural modeling. In particular, we argue that our current evaluation practices do not sufficiently account for semiotic complexity. 

We choose the phrase semiotic complexity partly to highlight the ways in which our modeling tools are not fundamentally linguistic. The machinery which is useful in modeling culture may just as well apply to very different kinds of problems, such as protein folding \cite{offert2024synthesizingproteinsgraphicscard}.
 
To more precisely define what we mean, we will now unpack our own theoretical commitments. We view the problem through the lens of Peircean semiotics, following its use by several others (e.g., \cite{ciula2017modelling, yadav2025culturalevaluationsvisionlanguagemodels, weatherby_justie_2022}). Peirce defines a sign as something that stands for something to someone \cite{Peirce_1931}. In other words, the mapping from a sign's representation to its meaning can only be resolved in the context of a perspective—the someone to whom the sign stands. Semiotic complexity, then, can be understood as the degree of perspectival variance a sign enables; i.e., the extent to which what a sign stands for may vary as a function of varying to whom the sign stands. In the cultural domain, a text like \textit{Zhuangzi} or the boundary between religion and irreligion exhibit much greater semiotic complexity than a word of greeting or the boundary between orthographically distinct languages. 
   
In the Peircean framework, signs may be involved in three kinds of reference: iconic, indexical, and symbolic. We follow aspects of Deacon's interpretation of Peirce \cite{Deacon_1998}, especially as it is drawn on in the ethnographic work of Kohn \cite{Kohn_2013}, assuming that symbolic reference allows for emergent semiotic phenomena that, while ultimately grounded in the lower-order substrate of iconic and indexical reference, cannot be neatly reduced to them. In the computational humanities, our data can typically be understood as artifacts of symbolic reference. But the data modality itself (e.g., language data) is insufficient for establishing the degree of semiotic complexity that is present. For example, the boundary between religious and irreligious text entails much greater semiotic complexity than the boundary between English and Mandarin. The latter boundary can be grounded in iconic reference: A given token will bear resemblance to other tokens from one of the orthographically distinct languages. The boundary between religion and irreligion is not so straightforwardly grounded in non-symbolic reference.

Similarly, the problem of translating \textit{Zhuangzi} into English entails greater semiotic complexity than a translation of a greeting. It is not the case that translating it can be decomposed into smaller translation problems that are semiotically less complex, each reducible to something like translating a word of greeting. Yet the dominant modeling practices commit this sort of reductionism, often by attempts to justify one set of modeling decisions as optimal, committing to a theory that optimal interpretations or translations are ontologically real. Such reductionism risks losing what makes our objects of study cultural in the first place. If the semiotic complexity of our work is not adequately grappled with, we risk our models giving us answers to questions that are unknown to us.

\section{Recommendations}
\label{sec:recommendations}
As a starting point for dealing with the semiotic complexity underlying much of our collective work, we make three recommendations for researchers. 

\paragraph{Recommendation 1.} We should engage in and expect articulations of our translation theories to enable greater interpretive and evaluative clarity, and ward against Nguyen's seduction of clarity. We should value this in the same way we value the source code used in an analysis.

\paragraph{Recommendation 2} We should hold our methods, especially our evaluation methods, more at arm's length, considering them as particular interpretive choices out of many other possible choices. It is not the case that model evaluation and performance metrics have no place in contexts of high semiotic complexity, but rather that evaluation methods cannot justify themselves. We can do this by contextualizing our methods in translation theories, but also by favoring methodological pluralism as necessary for mirroring the interpretive pluralism inherent in our research questions about culture.

We should then turn to consider what kinds of modeling practices would preserve the semiotic complexity of culturally-emergent research questions. If our research questions concern a complex semiotic system, like human conceptions of religion, then we ought to recognize the value in maximizing the interpretive variance of our models for answering such questions. In other words, complex questions require complex models, not in the sense of one large model, which cannot be semiotically complex given that it instantiates one particular interpretive lens. Rather, we should consider the use of multiple internally coherent, though mutually incoherent, models as the basis for an interpretive ecology that preserves the semiotic complexity of the cultural objects we are attempting to analyze. In a sense, we are arguing for a kind of computational thick description for our research questions as discussed in \cite{kommers2025meaningmetricusingllms}. 

Modeling decisions, performance measures, etc., function as signs themselves \cite{ciula2017modelling}. As complex signs, we ought to understand them relationally, rather than in the ordinal fashion that supervised evaluation imposes. In other words, it is not enough for a model to maximize one of many possible evaluation metrics. Instead, we would need to relate the interpretive constraints entailed by optimizing for that metric by relating it to others so that their individual interpretive consequences may come into view. This can also be extended to other modeling decisions such as model types and complexity. Grounding a complex language model such as a transformer-based neural network in a clear theoretical framework that connects it to culture and meaning will likely only be realistic once we have done it for simpler linear models. 

This paper itself presents the first theoretical stage in an attempt to construct such a framework for modeling practices.

\paragraph{Recommendation 3.} We recommend viewing the epistemological implications of semiotic complexity as being parallel to other problems in frontier sciences. By picking up the tools of science---and therefore, their distinct epistemological affordances---the field is necessarily engaged in science, and should take that seriously by learning from and contributing to other scientific fields. The problems posed by semiotic complexity are not wholly novel. Instead, issues of emergence, complexity, and reductionism lie at the heart of various kinds of frontier sciences, loosely grouped under the umbrella term of complexity science (see \cite{Mitchell_2009, Krakauer_FP_intro_2024}).
We argue that computational analyses of culture constitute a particular kind of complex system that will benefit from engagement with complex systems research while also being poised to make important contributions. 

Additionally, similar epistemological tensions around modeling practices are instructive for us, such as the use of p-values \cite{Gelman_Stern_2006}, the need to evaluate evaluative practices themselves \cite{Bak-Coleman_Devezer_2024}, and the critical role played by theory \cite{Guest_2024, Devezer_2024}.

\section{Conclusion}
\label{sec:conclusions}

We have argued the computational humanities are concerned with emergent cultural phenomena that exhibit high semiotic complexity, but that our dominant modeling practices do not adequately account for this complexity. As a result, we risk making significant epistemological translation errors by engaging in modeling practices that are odds with our own theoretical commitments. We have argued that, just as the evaluation and interpretation of translation work is aided by the articulation of the translator's theory for their work, so too is the translation work done in this field from the cultural to the computational. Without theorizing, we risk losing sight of the interpretive nature of our modeling efforts, blinded by the superficial clarity of narrow performance metrics. In laying out our own theoretical commitments underlying our articulation of semiotic complexity, we have also argued that complex semiotic phenomena demand pluralistic modeling approaches and broader engagement with other frontier sciences beyond computer science.

\section*{Acknowledgments}

ZS acknowledges the support of NSF RII Track-1: Data Analytics that are Robust and Trusted (DART): From Smart Curation to Socially Aware Decision Making, under Award Number OIA-1946391. The views expressed in this work are those of the authors and do not necessarily reflect the views of the agency.

\bibliographystyle{plain}
\bibliography{bibliography}

\end{document}